\documentclass[conference]{IEEEtran}
\IEEEoverridecommandlockouts
\usepackage{cite}
\usepackage{amsmath,amssymb,amsfonts,amsthm,balance}
\usepackage{algorithmic,hyperref}
\usepackage{graphicx}
\usepackage{textcomp}
\usepackage{xcolor}
\def\BibTeX{{\rm B\kern-.05em{\sc i\kern-.025em b}\kern-.08em
    T\kern-.1667em\lower.7ex\hbox{E}\kern-.125emX}}

\newcommand{\bA}        {\mathbf{A}}

\newcommand{\bb}        {\mathbf{b}}
\newcommand{\bB}        {\mathbf{B}}
\newcommand{\cB}        {\mathcal{B}}

\newcommand{\tnsrB}     {\underline{\bB}}

\newcommand{\bC}        {\mathbf{C}}

\newcommand{\E}         {\mathbb{E}}

\newcommand{\bG}        {\mathbf{G}}

\newcommand{\tnsrG}     {\underline{\bG}}

\newcommand{\cL}        {\mathcal{L}}

\newcommand{\cN}        {\mathcal{N}}

\newcommand{\cO}        {\mathcal{O}}

\newcommand{\bbP}       {\mathbb{P}}

\newcommand{\cP}        {\mathcal{P}}

\newcommand{\bQ}        {\mathbf{Q}}

\newcommand{\R}         {\mathbb{R}}

\newcommand{\bs}        {\mathbf{s}}

\newcommand{\bU}        {\mathbf{U}}

\newcommand{\bx}        {\mathbf{x}}
\newcommand{\bX}        {\mathbf{X}}

\newcommand{\tnsrX}     {\underline{\bX}}

\newcommand{\by}        {\mathbf{y}}

\DeclareMathOperator*{\argmin}  {arg\,min}                              

\newcommand{\diag}      {\mathrm{diag}}                                 
\newcommand{\rank}      {\mathrm{rank}}                                 
\newcommand{\tvec}      {\mathrm{vec}}                                  

\newcommand{\mi}[2]{\mathbb{I}(#1; #2)}
\newcommand{\cmi}[3]{\mathbb{I}(#1; #2 | #3)}

\newcommand{\norm}[1]{ \left\| #1 \right\| }

\newcommand{\Bvl}{\mathbf{B}_{l}}
\newcommand{\Bvlprime}{\mathbf{B}_{l'}}

\newcommand{\sll}{\mathbf{s}_l}
\newcommand{\slprime}{\mathbf{s}_{l'}}
\newcommand{\sff}{\mathbf{s}_f}
\newcommand{\sfprime}{\mathbf{s}_{f'}}

\newcommand{\Sp}{\mathbf{S}_{p}}
\newcommand{\Spprime}{\mathbf{S}_{p'}}
\newcommand{\Spbone}{\mathbf{S}_{p_1}}
\newcommand{\Spboneprime}{\mathbf{S}_{p_1'}}
\newcommand{\Spbtwo}{\mathbf{S}_{p_2}}
\newcommand{\Spbtwoprime}{\mathbf{S}_{p_2'}}

\newcommand{\spbone}{\mathbf{s}_{p_1,j}}
\newcommand{\spbtwo}{\mathbf{s}_{p_2,j}}

\newcommand{\gfone}{\mathbf{g}_{f}^{(1)}}

\newcommand{\Boneone}{\mathbf{B}_{1,p_1}^{(1)}}

\newcommand{\Btwoone}{\mathbf{B}_{2,p_2}^{(1)}}

\newcommand{\boneone}{\mathbf{b}_{1,p_1,j}^{(1)}}
\newcommand{\btwoone}{\mathbf{b}_{2,p_2,j}^{(1)}}
\newcommand{\gfonei}{g_{f,i}^{(1)}}

\newcommand{\Bonetwo}{\mathbf{B}_{1,p_1}^{(2)}}

\newcommand{\Btwotwo}{\mathbf{B}_{2,p_2}^{(2)}}

\newcommand{\bonetwo}{\mathbf{b}_{1,p_1,j}^{(2)}}

\newcommand{\gftwo}{\mathbf{g}_{f}^{(2)}}
\newcommand{\gftwoprime}{\mathbf{g}_{f'}^{(2)}}
\newcommand{\gftwoi}{g_{f,i}^{(2)}}

\newcommand{\Bpone}{\mathbf{B}_{1,p_1}}
\newcommand{\Bponeprime}{\mathbf{B}_{1,p'_1}}
\newcommand{\Bptwo}{\mathbf{B}_{1,p_2}}
\newcommand{\Bptwoprime}{\mathbf{B}_{1,p'_2}}
\newcommand{\gf}{\mathbf{g}_{f}}
\newcommand{\gfprime}{\mathbf{g}_{f'}}
\theoremstyle{plain}
\newtheorem{lemma}{Lemma}
\newtheorem{theorem}{Theorem}
\newtheorem{corollary}{Corollary}

\theoremstyle{definition}

\theoremstyle{remark}

\begin{document}

\title{A Minimax Lower Bound for Low-Rank Matrix-Variate Logistic Regression}%

\author{\IEEEauthorblockN{Batoul Taki, Mohsen Ghassemi, Anand D. Sarwate, and Waheed U. Bajwa}
\IEEEauthorblockA{Department of Electrical and Computer Engineering, Rutgers University--New Brunswick, Piscataway, NJ}\\
E-mail:{\tt\{batoul.taki, mohsen.ghassemi, anand.sarwate, waheed.bajwa\}@rutgers.edu}
\thanks{This work was supported by the US NSF Grants CCF-1910110 and CCF-1453073.}
}

\maketitle

\begin{abstract}
   This paper considers the problem of matrix-variate logistic regression. It derives the fundamental error threshold on estimating low-rank coefficient matrices in the logistic regression problem by obtaining a lower bound on the minimax risk. The bound depends explicitly on the dimension and distribution of the covariates, the rank and energy of the coefficient matrix, and the number of samples. The resulting bound is proportional to the intrinsic degrees of freedom in the problem, which suggests the sample complexity of the low-rank matrix logistic regression problem can be lower than that for vectorized logistic regression. \color{red}\color{black} The proof techniques utilized in this work also set the stage for development of minimax lower bounds for tensor-variate logistic regression problems.
\end{abstract}


\section{Introduction}
Logistic Regression (LR) is a statistical model used commonly in machine learning for classification problems \cite{wright1995logistic}. In the simplest terms, LR seeks to model the conditional probability that a categorical response variable $y_i$ takes on one of two possible values, $y_i \in \{0,1\}$, which represents one of two possible events taking place (such as success/failure and detection/no detection). In regression analysis, the aim is to accurately estimate the model class parameters through a set of training data points $\{ \bx_i, y_i\}_{i = 1}^{n}$ , where $\bx_i$ is the $i^{th}$ covariate sample vector. The conventional LR model is as follows:
\begin{align}
    \bbP_{y|\bx}(y_i = 1 |\bx_i) = \frac{1}{1+\exp^{-(\bb^T\bx_i + z)}}, \label{vector_model}
\end{align}
where $y_i \in \{ 0,1 \}$ is the binary response, $\bx_i \in \R^{m} $ is the known covariate vector, $\bb \in \R^{m}$ is the deterministic but unknown coefficient vector, and $z$ is the intercept where $\E[z] = 0$. 

In many practical applications, covariates naturally take the form of two-dimensional arrays. Common examples include images, biological data such as electroencephalography (EEG), fiber bundle imaging \cite{dumas2019computational} and spatio-temporal data \cite{lodhi2020learning}. Classical machine learning techniques vectorize such matrix covariates and estimate a coefficient vector. This leads to computational inefficiency and the destruction of the underlying spatial structure of the coefficient matrix, which often carries valuable information~\cite{hung2013matrix,zhang2018rank}. To address these issues, one can model the matrix LR problem as
\begin{align}
    \bbP_{y|\bX}(y_i = 1 | \bX_i) = \frac{1}{1+\exp^{-(\langle\bB,\bX_i\rangle + z)}}, \label{gen_matrix_model}
\end{align}
where $\bX_i \in \R^{m_1\times m_2}$ is the known covariate matrix, $\bB \in \R^{m_1 \times m_2}$ is the coefficient matrix, and $z$ is the intercept. The model in \eqref{gen_matrix_model} preserves the matrix structure of, and the valuable information in, the covariate data. In matrix-variate logistic regression analysis, the goal is to find an estimate of the coefficient $\bB$. 

In this paper, we focus on the high-dimensional setting where $n \ll m_1m_2$ and derive a lower bound on the minimax risk of the matrix LR problem under the assumption that $\bB$ has rank $r \ll \min\{m_1, m_2\}$. We note that the low-rank assumption has been used ubiquitously in regression analysis in former works \cite{barnes2019minimax, zhang2020islet}. This low-rank structure may arise from the presence of latent variables, such that the model's intrinsic degrees of freedom are smaller than its extrinsic dimensionality. This allows us to represent the data in a lower-dimensional space and potentially reduce the sample complexity of estimating the parameters. Minimax lower bounds are useful in understanding the fundamental error thresholds of the problem under study, and in assessing the performance of the existing algorithms. They also provide beneficial insights to the parameters on which an achievable minimax risk might depend.

Whilst a number of works in the literature derive minimax lower bounds for higher-order linear regression problems \cite{raskutti2019convex, zhang2020islet}, as well as vector LR problems \cite{barnes2019minimax, foster2018logistic, abramovich2018high}, to the best of our knowledge, little work has been done on this topic in matrix-variate LR problems. Regarding the existing literature, there are several works that study the matrix LR problem. Many works extend the matrix LR problem of \eqref{gen_matrix_model} by proposing regularized matrix LR models in order to obtain rank-optimized or sparse estimates of the coefficient matrix in the model in \eqref{gen_matrix_model} \cite{zhang2018rank, shi2014sparse}. Recently, works have introduced regularized matrix LR models for inference on image data \cite{an2020logistic}. Much of the literature develops efficient algorithms and provides empirical results on their performance \cite{zhang2018rank, shi2014sparse, an2020logistic} while some provide theoretical guarantees for the proposed algorithms \cite{shi2014sparse}. By contrast, Baldin and Berthet \cite{berthet2020statistical} model the problem of graph regression as a matrix LR problem, where $\bB \in \R^{m\times m}$. The matrix $\bB$ is assumed to be block-sparse, low-rank, and square. Under these assumptions (which are restrictive and not the most general), the authors derive minimax lower bounds on the estimation error $\norm{\widehat{\bB} - \bB}_F^2$.

\color{red}\color{black}The three prior works \cite{hung2013matrix, shi2014sparse, an2020logistic} implement algorithms for matrix logistic regression but do not prove sample complexity bounds (upper or lower). \color{black} In this paper, we derive a minimax lower bound on the error of a low-rank LR model which gives a bound on the number of samples necessary for estimating $\bB$. Contrary to prior works, we impose minimal assumptions on $\bB$, making our results generalizable to a larger class of matrices. 

Our Minimax lower bound relies on the distribution of the covariates and the energy of the regression matrix. Following previous works that derive minimax lower bounds in the dictionary learning setting \cite{jung2016minimax, shakeri2016minimax}, we use the standard strategy of lower bounding the minimax risk in parametric estimation problems by the maximum error probability in a multiple hypothesis test problem and using Fano's inequality \cite{yu1997assouad}. We derive a lower bound that is proportional to the degrees of freedom in the rank-$r$ matrix LR problem, and we reduce the sample complexity from $\cO(m_1m_2)$ in the vector setting to $\cO(r(m_1 + m_2))$. \color{red} \color{black}This result is intuitively pleasing as it coincides with the number of free parameters in the model. We also show that our methods are easily extendable to the tensor case, i.e., $\tnsrB$ is a coefficient tensor with some known properties.


\section{Model and Problem Formulation}\label{model_prob_form}

We use the following notation convention throughout the paper: $x$, $\bx$, $\bX$ and $\tnsrX$ denote scalars, vectors, matrices and tensors, respectively. We denote $\mathbf{I}_m$ as the $m \times m$ identity matrix. Also, $\norm{\bx}_0$, $\norm{\bx}_2$ and $\norm{\bX}_F$ denote the $\ell_0$, $\ell_2$ and Frobenius norms, respectively. Given a matrix $\bX$, $\bx_j$ is the $j^{th}$ column of $\bX$.
If $\tnsrX$ is a third-order tensor, then by fixing indices in the first and second modes, the matrix $\tnsrX_{:,:,j}$ is the $j^{th}$ frontal slice of $\tnsrX$.
If $a \in \R$, then $\lfloor a \rfloor$ is the greatest integer less than $a$. We denote $\bA \otimes \bC$ as the Kronecker product of matrices $\bA$ and $\bC$. We define the function $\tvec(\bX)$ as the column-wise vectorization of matrix $\bX$. Finally, $[R]$ is shorthand for $\{1, \dots, R\}$. 

Consider the matrix LR model in \eqref{gen_matrix_model}, in which the Bernoulli response $y_i \in \{ 0 ,1 \}$ is the $i^{th}$ response variable of $n$ independent and identically distributed (i.i.d) observations. The covariate matrix $\bX_i \in \R^{m_1 \times m_2}$ has independent, normally distributed, zero-mean and $\sigma^2$ variance entries. The response $y_i$ is generated according to \eqref{gen_matrix_model} from $\bX_i$ and a fixed coefficient matrix $\bB \in \R^{m_1 \times m_2}$ with $\norm{\bB}_F^2$ upper-bounded by a constant.   

We consider the case where $\bB $ is a rank-$r$ matrix. Specifically, the rank-$r$ singular value decomposition of $\bB$ is
\begin{align}
    \bB = \bB_1 \bG \bB_2^T, \label{B_matrix}
\end{align}
where $\bB_1 \in \R^{m_1 \times r}$ and $\bB_2 \in \R^{m_2 \times r}$ are (tall) singular vector matrices with orthonormal columns, and $\bG = \diag(\lambda_1, \cdots, \lambda_r) \in \R^{r \times r}$ is the matrix of singular values with $\lambda_i > 0, \forall i \in [r]$. Under this low-rank structure of $\bB$, we have
\begin{align}
    \bbP_{y|\bX}(y_i = 1|\bX_i) = \frac{1}{1+ \exp^{-(\langle\bB_1\bG\bB_2^T,\bX_i\rangle +z)}}. \label{rank_matrix_model}
\end{align}
We use Kronecker product properties \cite{kolda2009tensor} to express $\bb = \tvec(\bB)$ as $\bb = \left(\bB_2 \otimes \bB_1\right)\tvec(\bG)$.

The goal in LR is to find an estimate $\widehat{\bB}$ of $\bB$ using the training data $\bigl\{\bX_i, y_i\bigr\}_{i=1}^{n}$. We assume that the true coefficient matrix $\bB$ belongs to the set
    \begin{align}
    \cB_d(\mathbf{0}) \triangleq \{\bB' \in \cP_r : \rho(\bB',\mathbf{0}) < d \}, \label{open_ball}
    \end{align}
the open ball of radius $d$ with distance metric $\rho = \norm{\cdot}_F$, which resides in a given the parameter space, 
\begin{align}
    \nonumber \cP_r \triangleq & \biggl\{ \bB' \in \R^{m_1 \times m_2}: \rank(\bB') = r, ~\bB' = \bB'_1 \bG' \bB'_2, \\ \nonumber &\norm{\bb'_{1,j}}_2 = \norm{\bb'_{2,j}}_2 =1, \bb'_{k,j} \perp \bb'_{k,j'},  \forall j,j' \in [r], \\ & j \neq j' , k \in \{1,2\}\biggr\} ~\cup ~\{ \mathbf{0} \}. \label{param_space}
\end{align}
Thus $\cB_d(\mathbf{0}) \subset \cP_r$ and $\bB$ has energy bounded by $\norm{\bB}_F^2< d^2$. 


Our objective is to find a lower bound on the minimax risk for estimating coefficient matrix $\bB$. We define the non-decreasing function $\phi =\norm{\cdot}_F^2  \triangleq \R_+ \rightarrow \R_+$ with $\phi(0) = 0$. The minimax risk is thus defined as the worst-case mean squared error (MSE) for the best estimator, i.e., 
\begin{align}
    \varepsilon^* = \underset{\widehat{\bB}}{\inf} \underset{\bB \in 
    \cB_d(\mathbf{0})}{\sup} \E_{\by,\tnsrX^c} \biggl\{\norm{\widehat{\bB} - \bB}_F^2 \biggr\}, \label{mm}
\end{align}
where $\tnsrX^c$ is the covariate tensor of $n$ samples. We point out that minimax risk in \eqref{mm} is an inherent property of the LR problem and holds for all possible estimators. 



\section{Minimax Lower Bound For Low-Rank Matrix Logistic Regression}\label{Minimax_Matrix}
The minimax lower bounds in the low-rank matrix LR setting that are based on Fano's inequality will depend explicitly on the dimensions of the covariate matrices $\bX_i$ and their distribution, the rank, the upper bound on the energy of the coefficient matrix $\bB$, the number of samples, $n$, and the construction of the multiple hypothesis test set.

The novelty of our work is that we explicitly leverage the rank-$r$ structure of the coefficient matrix $\bB$, in \eqref{B_matrix}, which leads to the construction of each hypothesis that is structurally different in our setting compared to vector LR \cite{barnes2019minimax, abramovich2018high, foster2018logistic}. \color{red}\color{black} Furthermore, existing low-rank matrix packings, such as those in Negahban and Wainwright~\cite{negahban2012restricted} are not useful in the LR setting of this paper, since the logistic function in our model makes part of our analysis non-trivial and fundamentally different to such works. \color{black}\color{black}In this work, we create a set $\cB_L$ from three constructed sets (for $\bB_1$, $\bB_2$ and $\bG$), and derive conditions under which all sets can exist. These conditions ensure the existence of a hypothesis set, $\cB_L$, with elements that have a rank-$r$ matrix structure and additional essential properties noted below. \color{red}\color{black} We also show that our chosen constructed hypothesis set and analysis are more suitable for our matrix-variate model because they are easily generalizable to the tensor setting for LR.


\subsection{Main Result}
\color{black} We derive lower bounds on $\varepsilon^*$ using an argument based on Fano's Inequality~\cite{khas1979lower}. To do so, we first relate the minimax risk in \eqref{mm} to a multi-way hypothesis testing problem between an exponentially large family, with respect to the dimensions of the matrices, of $L$ distinct coefficient matrices with energy upper bounded by $d^2$, i.e., matrices residing inside the openball $\cB_d(\mathbf{0})$ of fixed radius:
\begin{align}
    \cB_L = \{ \Bvl \colon l \in [L]\} \subset \cB_d(\mathbf{0}), \label{hyp_set}
\end{align}
where the correct hypothesis $\Bvl$ is generated uniformly at random from the set $\cB_L$.

More specifically, suppose there exists an estimator with worst-case MSE that matches $\varepsilon^*$. This estimator can be used to solve a multiple hypothesis testing problem using a minimum distance decoder. The minimax risk can then be lower bounded by the probability of error in such a multiple hypothesis test. Our main challenge is to further lower bound this probability of error in order to derive lower bounds on the minimax risk.

We now state our main result on the minimax risk of the low-rank matrix coefficient estimation problem in LR.

\begin{theorem}\label{matrix_mm_LB}
    Consider the rank-$r$ matrix LR problem in \eqref{rank_matrix_model} with $n$ i.i.d observations, $\bigl\{ \bX_i, y_i\bigr\}_{i=1}^{n}$ where the true coefficient matrix $\norm{\bB}_F^2 <d^2$. 
    Then, for covariate $\tvec(\bX_i) \sim \cN(0,\sigma^2 \mathbf{I}_m)$, the minimax risk is lower bounded by
    \begin{align}
       \varepsilon^* \geq \frac{\biggl(\biggl[c_2\bigl(c_1r(m_1 + m_2 -2) +  c_1(r-1)\bigr) 
        - c_3 \biggr] -1\biggr)}{8 n\sigma\sqrt{\frac{2}{\pi}}} , \label{mm_LB}
    \end{align}
    where $c_1 = \left(1 -\frac{1}{10}\right)^2$, $c_2 = \frac{\log_2(e)(\sqrt{2}-1)}{4\sqrt{2}}$ and $c_3 =(\frac{3(\sqrt{2} -1)}{\sqrt{8}})\log_2\left(\frac{3}{2}\right)$.
\end{theorem}
We discuss the implications of \eqref{mm_LB} of Theorem~\ref{matrix_mm_LB} in Section~\ref{disc_conc}

\subsection{Roadmap for Theorem \ref{matrix_mm_LB}}\label{roadmap}
\color{black} For notational convenience, we define the response vector of $n$ samples as $\by = [y_1, \cdots, y_n]$, and the covariate tensor of $n$ samples as $\tnsrX^c \in 
R^{m_1 \times m_2 \times n}$, where the $i^{th}$ frontal slice, $\tnsrX_{:,:,i}^c$, is the covariate matrix $\bX_i$ of the $i^{th}$ sample. 

For our analysis, we construct a hypothesis set of $L$ distinct matrices, as defined in \eqref{hyp_set}. However, each hypothesis $\Bvl$ is a low-rank matrix following the decomposition in \eqref{B_matrix} and is composed of three separately constructed sets (for $\bB_1$, $\bB_2$, and $\bG$). Now, consider $\mi{\by}{l}$, the mutual information between the observations $\by$ and random index $l$. The construction of the set $\cB_L$ is used to provide upper and lower bounds on $\mi{\by}{l}$, from which we derive lower bounds on the minimax risk $\varepsilon^*$.

For finding a lower bound on $\mi{\by}{l}$, we first consider the exponentially large packing set $\cB_L$, with respect to the dimensions of $\bB$, where any two distinct hypotheses $\Bvl$ and $\Bvlprime$ are separated by a minimum distance, i.e.,
\begin{align}
    \norm{\Bvl - \Bvlprime}_F^2 \geq 8\delta \label{pack_dist}
\end{align}
for some positive $\delta$. To achieve our desired bounds on the minimax risk, we require the existence of an estimator producing estimate $\widehat{\bB}$ and achieving the minimax bound $\varepsilon^* < \sqrt{\delta}$. We consider the minimum distance decoder 
\begin{align}
    \widehat{l}(\by) \triangleq \underset{\Bvlprime \in \cB_d(\mathbf{0})}{\argmin} \norm{\widehat{\bB} - \Bvlprime}_F^2 \label{min_dist_dec},
\end{align}
which seeks to detect the correct coefficient matrix $\Bvl$. We require the estimate $\widehat{\bB}$ to satisfy $\norm{\widehat{\bB} - \Bvl}_F^2 <\sqrt{2\delta}$, for the minimum distance decoder to detect $\Bvl$ and for the probability of detection error to be $\bbP(\widehat{l}(\by) \neq l) = 0$. A detection error might occur when $\norm{\widehat{\bB} - \Bvl}_F^2 \geq \sqrt{2\delta}$. The following bound relates the loss $\norm{\widehat{\bB} - \Bvl}_F^2$ and the probability of error $\bbP(\widehat{l}(\by) \neq l)$:
\begin{align}
    \bbP(\widehat{l}(\by) \neq l) \leq \bbP\Big(\norm{\widehat{\bB} - \Bvl}_F^2 \geq \sqrt{2\delta}\Big). \label{relate_prob_error}
\end{align}
Next, \eqref{relate_prob_error} is used to obtain a lower bound on $\mi{\by}{l}$ using Fano's inequality stated below \cite{yu1997assouad}:
\begin{align}
   \mi{\by}{l} \geq \left(1-\bbP(\widehat{l}(\by) \neq l)\right) \log_2(L)-1 \triangleq u_1. \label{fano}
\end{align}

Secondly, for finding an upper bound on $\mi{\by}{l}$, we recognize that the LR model produces response variables $y_i$ that are Bernoulli random variables conditioned on $\bX_i$. On the basis thereof, we evaluate the mutual information, conditioned on $\tnsrX^c$, $\cmi{\by}{l}{\tnsrX^c}$. Next, define $f_l(\by|\tnsrX^c)$ as the conditional probability distribution of $\by$ given $\tnsrX^c$ with coefficient matrix $\Bvl$, and let $D_{KL}$ be the relative entropy, between any two $f_l(\by|\tnsrX^c)$ and  $f_{l'}(\by|\tnsrX^c)$, produced from two distinct $\Bvl$, $\Bvlprime$. Due to the convexity of $-\log$, we upper bound $D_{KL}$ as follows \cite{cover2012elements, jung2016minimax}:
\begin{align}
    \cmi{\by}{l}{\tnsrX^c} \leq \frac{1}{L^2} \sum_{l,l'}\E_{\tnsrX^c} D_{KL}(f_l(\by|\bX)||f_{l'}(\by|\bX)) \triangleq u_2\label{kl}.
\end{align} 

Lastly, we remark that from \cite{jung2016minimax} we have $\mi{\by}{l} \leq \cmi{\by}{l}{\tnsrX^c}$, and $\cmi{\by}{l}{\tnsrX^c}$ is trivially lower bounded by \eqref{fano}. Thus, \eqref{fano} and \eqref{kl} give rise to upper and lower bounds on the conditional mutual information:
\begin{align}
    u_1 \leq \cmi{\by}{l}{\bX} \leq u_2, \label{cmi_dual}
\end{align}
where $u_1$ and $u_2$ require evaluation, and from which we obtain a lower bound on the minimax risk.  
\section{Proof of Main Result}\label{proof}
The formal proof for Theorem \ref{matrix_mm_LB} relies on three lemmas. Lemma \ref{vector_set} introduces the exponentially large, with respect to the dimensions, set of vectors from which we will later construct $\bG$. The set is constructed as a subset of an $(r-1)$-dimensional hypercube with a required minimum distance between any two distinct elements in the set. Lemma \ref{vector_set} bounds the probability that this minimum distance is violated.
\begin{lemma}\label{vector_set}
    Let $r >0$ and $F \geq 2$. Consider the set of $F$ vectors $\bigl\{ \sff \in \R^{r-1} : f \in [F]  \bigr\}$, where each entry in vector $\sff$ is an independent and identically distributed random variable taking values $\biggl\{-\frac{1}{\sqrt{r -1}}, +\frac{1}{ \sqrt{r -1}}\biggr\}$ uniformly. The probability that there exists a distinct pair $(f, f')$ such that $\norm{\sff- \sfprime}_0 < \frac{r-1}{20}$ is upper bounded as follows:
    \begin{align}
        & \nonumber \bbP(\exists (f,f') \in [F] \times [F], f\neq f' : \norm{\sff- \sfprime}_0 < \frac{r-1}{20} )\\
        & \leq \exp\left[2\log(F) - \log(2) - \frac{1}{2}\left(1 - \frac{1}{10}\right)^2(r-1)\right].
    \end{align}
\end{lemma}
The above is a direct result of a standard application of McDiarmid's inequality~\cite{dubhashi2009concentration}. Notice, however, that our technique differs from \cite{shakeri2018minimax} because we are imposing minimum distance conditions in the Hamming metric, rather than conditions on the inner product, between vectors. 

Similar to the above, the following corollary introduces a set of matrices from which we will construct $\bB_1, \bB_2$.   

\begin{corollary}\label{matrix_set}
    Let $m, r >0$ and $P \geq 2$. Consider the set of $P$ matrices $\bigl\{ \Sp \in \R^{(m-1) \times r }: p \in [P]  \bigr\}$, where each entry in matrix $\Sp$ is an independent and identically distributed random variable taking values $\biggl\{-\frac{1}{\sqrt{(m -1)r}}, +\frac{1}{ \sqrt{(m-1)r}} \biggr\}$ uniformly. The probability that there exists a distinct pair $(p, p')$ such that $\norm{\Sp- \Spprime}_0 \leq \frac{(m-1)r}{20}$ is upper bounded as follows:
    \begin{align}
        & \nonumber \bbP(\exists (p,p') \in [P] \times [P], p\neq p' : \norm{\Sp- \Spprime}_0 < \frac{(m-1)r}{20} )\\
        & \leq \exp\left[2\log(P) - \log(2) - \frac{1}{2}\left(1 - \frac{1}{10}\right)^2(m-1)r\right].
    \end{align}

\end{corollary}

From the results in Lemma \ref{vector_set} and Corollary \ref{matrix_set}, Lemma \ref{packing_lemma} derives conditions on $L$ such that $\cB_L$ with a certain set of properties exists, and constructs two sets of matrices: 1) $m_1 \times r$ and $m_2 \times r$ orthonormal matrices from the set of \textquotedblleft generating" matrices defined in Corollary \ref{matrix_set}, and 2) $r \times r$ diagonal matrices from the set of \textquotedblleft generating" vectors in Lemma \ref{vector_set}. Each element $\Bvl \in \cB_L$ is constructed from the three sets defined above, according to the decomposition in \eqref{B_matrix}. Next, we prove lower and upper bounds on the distance between any two distinct elements in $\cB_L$. The lower bound determines the minimum packing distance between any two $\Bvl, \Bvlprime$, whilst the upper bound is used to derive the results in Lemma \ref{cmi_UB_lemma}.
\begin{lemma}\label{packing_lemma}
    There exists a collection of $L$ matrices $B_L \triangleq \{\Bvl: l \in [L]\} \subset \cB_d(\mathbf{0})$ for some $d> 0$ of cardinality 
    \begin{align}
        L = 2^{\lfloor \frac{\log_2(e)}{4}\left(\left(1 - \frac{1}{10}\right)^2\left(r(m_1 +m_2 -1\right) +  \left(1 - \frac{1}{10}\right)^2(r-1)\right) - \frac{3}{2}\log_2\left(\frac{3}{2}\right)\rfloor} \label{L_bound}
    \end{align} 
    such that for any
    \begin{align}
        \sqrt{\frac{8(r-1)}{r}} < \varepsilon \leq d\sqrt{\frac{r-1}{r}}, \label{eps_bounds}
    \end{align}
    we have
    \begin{align}
        \frac{r\varepsilon^2}{r-1}< \norm{\Bvl - \Bvlprime}_F^2 \leq 4\frac{r\varepsilon^2}{r-1}. \label{distance_bounds}
    \end{align}
\end{lemma} 

Lemma \ref{cmi_UB_lemma} derives an upper bound on $\cmi{\by}{l}{\tnsrX^c}$. 
\begin{lemma}\label{cmi_UB_lemma}
    Consider the matrix LR problem given by the model in \eqref{gen_matrix_model} such that $\bB \in \cB_d(\mathbf{0})$, for some $d >0$, and consider the set $\cB_L$ defined in Lemma \ref{packing_lemma}. Consider $n$ i.i.d observations $y_i$ following a Bernoulli distribution when conditioned on $\bX_i \sim 
    \cN(\mathbf{0}, \sigma^2 \mathbf{I})$, then we have,
    \begin{align}
        \cmi{\by}{l}{\bX} \leq n\frac{2}{r} \sqrt{\frac{2}{\pi}}\sigma \varepsilon. \label{cmi_UB}
    \end{align}
\end{lemma}

The final lemma bounds the error probability $\bbP(\widehat{l}(\by) \neq l)$ under the conditions, mentioned in Section \ref{roadmap}, on $\varepsilon^*$ and $\norm{\widehat{\bB} - \Bvl}_F^2$ for the recovery of hypothesis $\Bvl$ by the minimum distance decoder. The proof follows exactly that for Lemma 8 in \cite{jung2016minimax}.
\begin{lemma}\label{cmi_LB_lemma}
    Consider the minimum distance decoder in \eqref{min_dist_dec}, for the $L$ matrices constructed in Lemma \ref{packing_lemma}. Consider the LR regression model in \eqref{gen_matrix_model} with minimax risk $\varepsilon^*$, which is assumed to be upper bounded by $\varepsilon^* \leq \sqrt{\delta}$. The minimum distance decoder recovers the correct hypothesis if $\norm{\widehat{\bB} - \Bvl}_F^2 < \sqrt{2\delta}$. The detection error of the minimum distance decoder is upper bounded by $\bbP(\widehat{l}(\by) \neq l) \leq \frac{1}{\sqrt{2}}$.
\end{lemma}

\begin{IEEEproof}[Proof of Theorem \ref{matrix_mm_LB}]
Fix $r$, let $\varepsilon > 0$ satisfy \eqref{eps_bounds}, and define $\varepsilon^2 \triangleq \frac{8\delta(r-1)}{r}$. Suppose there exists an estimator $\widehat{\bB}$ which guarantees a risk $\varepsilon^* \leq \sqrt{\delta} = \sqrt{\frac{r}{8 (r-1)}}\varepsilon$. By Lemma~\ref{packing_lemma}, there exists a packing set $B_L$ containing $L$ distinct rank-$r$ matrices, where $L$ satisfies \eqref{L_bound} and for any pair $\Bvl,\Bvlprime \in B_L$, the distance $\norm{\Bvl - \Bvlprime}_F^2$ satisfies \eqref{pack_dist}. By Lemma \ref{cmi_UB_lemma}, the conditional mutual information $\cmi{\by}{l}{\tnsrX^c}$ is upper bounded by \eqref{cmi_UB}, and $\bbP(\widehat{l}(\by) - \Bvl)$ is upper bounded by Lemma~\ref{cmi_LB_lemma}. Replacing \eqref{cmi_UB} and \eqref{fano} in \eqref{cmi_dual}, we have:
\begin{align}
    \frac{\sqrt{2}-1}{\sqrt{2}}\log_2L -1 \leq \cmi{\by}{l}{\bX} \leq n\sigma\frac{2}{r}\sqrt{\frac{2}{\pi}}\varepsilon.
\end{align}
Lastly, if we set $\varepsilon^* =\sqrt{\frac{r}{8(r-1)}} \varepsilon$, then we achieve the result in \eqref{mm_LB}. 
\end{IEEEproof}


\section{Discussion and Conclusion}\label{disc_conc}
In this paper we provided a minimax lower bound on the low-rank matrix-variate LR problem in the high-dimensional setting. We constructed a packing set of low-rank structured matrices with finite energy. Using the construction, we derived bounds on the conditional mutual information defined in our problem, in order to obtain a lower bound on the minimax risk. 
Compared to the vector case, such as in \cite{barnes2019minimax}, the result in Theorem \ref{matrix_mm_LB} shows a decrease in the lower bound from $\cO(m_1m_2)$ to $\cO(r(m_1 +m_2)$). The result also shows that the lower bound on the minimax risk is proportional to the intrinsic degrees of freedom in the coefficient matrix LR, (i.e., $r(m_1 +m_2 + 1)$), and decreases with increasing sample size $n$. This suggests that we can develop algorithms that take advantage of the low-rank structure of the coefficient matrices. \color{red}\color{black} Moreover, the result in \eqref{mm_LB} can be generalized from low-rank matrices to low-rank tensors. Imposing a rank-$r$ decomposition on a coefficient tensor, $\tnsrB$, holds the same conveniences as those discussed above of low-rank matrices. This low-rank structure on $\tnsrB$ is a special case of the well-known Canonical Polyadic Decomposition or Parallel Factors (CANDECOMP/PARAFAC, or CP)\cite{kolda2009tensor}, formally defined as,
\begin{align}
    \tnsrB := \sum\limits_{h = 1}^{r} \lambda_h \bb_1^{(h)} \circ \cdots \circ \bb_K^{(h)}, \label{CP_model} 
\end{align}
where $\bb_k^{(h)} \in \R^{m_k}$ is a column vector along the $k^{th}$ mode of $\tnsrB$, $\circ$ is the outer product, and $\lambda_h \bb_1^{(h)} \circ \cdots \circ \bb_K^{(h)}$, for any rank $h$, is a rank-1 tensor weighted by $\lambda_h$. The rank-$r$ CP-decomposition expresses tensors as a sum of $r$ rank-1 tensors. Equivalent to \eqref{CP_model} is the following expression of a CP-structured tensor:
\begin{align}
    \tnsrB = \tnsrG \times_1 \bB_1 \times_2 \cdots \times_K \bB_K, \label{gen_CP_model}
\end{align}
where the tensor $\tnsrG \in \R^{\overbrace{r \times \cdots \times r}^{K\text{-times}}}$ is simply a higher-order analogue of the diagonal matrix $\bG$ in \eqref{B_matrix} (where only the elements along the super-diagonal of $\tnsrG$ are non-zero), and the mode-$k$ factor matrices $\bB_k \in \R^{m_k \times r}, ~\forall k \in [K]$ are rank-$r$ matrices with orthonormal columns. Thus, the setup and construction proposed in this paper can be extended to the tensor case, where $\bB$ is simply a special case of $\tnsrB$, with $K=2$ factor matrices. \color{black}

\section{Appendix}
\begin{IEEEproof}[Proof of Lemma 1]
Consider the set of $F$ vectors $\{\sff \in \R^{r-1} : f \in [F]\}$, in \eqref{vector_set}. For indices $f, f' \in [F]$, $f \neq f'$, define the vector $\widetilde{\bs} \triangleq \sff \odot \sfprime$ as the point-wise product of $\sll$ and $\slprime$, and $\widetilde{s}_i$ as the $i^{th}$ entry of $\widetilde{\bs}$, for $i \in [r-1]$. Define also the function
\begin{align}
    h(\widetilde{\bs}) \triangleq (r-1)\sum\limits_{i =1}^{r-1} \widetilde{\bs}_i.
\end{align}
We use $\widetilde{\bs} \approx \widetilde{\bs}'$ to say that $\widetilde{s}_i =  \widetilde{s}'_i$ for all entries $i \in [r-1]$, except one. 

We require a minimum packing distance 
\begin{align}
    \norm{\sff - \sfprime}_0 = \frac{1}{2}\left[(r-1) - (r-1)\sum\limits_{i =1}^{r-1} \widetilde{s}_i\right] > \frac{r-1}{20}. \label{distance_cond}
\end{align}
The probability that the requirement in \eqref{distance_cond} is not satisfied is defined as
\begin{align}
    \bbP\left((r-1)\sum_{i=1}^{r}\widetilde{\bs}_i \geq (r-1)( 1- \frac{1}{10} )\right).
\end{align}
The function $h(\cdot)$ satisfies
\begin{align}
     &\underset{\widetilde{\bs} \approx \widetilde{\bs}'}{\sup} \lvert h(\widetilde{\bs}) - h(\widetilde{\bs}')\rvert = \lvert (r-1)\frac{1}{r-1} + (r-1)\frac{1}{r-1} \rvert = 2. 
\end{align}
According to McDiarmid's inequality in \cite{dubhashi2009concentration}, for $\frac{(r-1)(9)}{10} > 0$, and since $\E_{\widetilde{\bs}} [h(\widetilde{\bs})] = 0$, we have,
\begin{align}
    \nonumber \bbP\biggl( (r-1)\sum_{i=1}^{r-1}\tilde{\bs}_i & \geq (r-1)\frac{9}{10}\biggr) \\ & \leq\exp\left[ -\frac{1}{2}\left(1-\frac{1}{10}\right)^2(r-1)\right] \label{proba_violated}.
\end{align}
The probability in \eqref{proba_violated} is for any two distinct pairs $(f, f') \in [F] \times [F]$. We take a union bound over all $\binom{L}{2}$ distinct pairs and upper bound the probability as:
\begin{align}
    & \nonumber \bbP(\exists (f,f') \in [F] \times [F], f\neq f' : (r-1)\sum_{i=1}^{r-1}\widetilde{s}_i \geq \frac{(r-1)9}{10} )\\
    &\nonumber \leq \frac{F^2}{2} \exp\left[- \frac{1}{2}\left(1-\frac{1}{10}\right)^2(r-1)\right]\\
    & = \exp\left[2\log(F) - \log(2) - \frac{1}{2}\left(1 - \frac{1}{10}\right)^2(r-1)\right] \label{proba_union_violated}.
\end{align}
\end{IEEEproof}

\begin{IEEEproof}[Proof of Lemma 2]
Fix the following arbitrary real orthogonal bases: $\bQ \text{ of } \R^{r}$, the set of distinct $r$ bases, $\bigl\{\bU_{1,j}\bigr\}_{j=1}^r \text{ of } \R^{m_1}$, and the set of distinct $r$ bases, $\bigl\{\bU_{2,j}\bigr\}_{j=1}^r \text{ of } \R^{m_2}$.

Next, consider the following hypercubes or subsets thereof: 1) The set of $F$ vectors $\{\sff\}$ from Lemma \ref{vector_set}: 
\begin{align}
    \sff \in \biggl\{\frac{-1}{\sqrt{r-1}}, \frac{+1}{\sqrt{r-1}}\biggr\} \subset \R^{r-1},\label{g_set}
\end{align}
2) Two sets of $P_1$ and $P_2$ matrices, from Corollary \ref{matrix_set}:
\begin{align}
    \Spbone \in \biggl\{\frac{-1}{\sqrt{(m_1-1)r}}, \frac{+1}{\sqrt{(m_1-1)r}}\biggr\} \subset \R^{(m_1-1)\times r},\label{b1_set}
\end{align}
and
\begin{align}
    \Spbtwo \in \biggl\{\frac{-1}{\sqrt{(m_2-1)r}}, \frac{+1}{\sqrt{(m_2-1)r}}\biggr\} \subset \R^{(m_2-1)\times r},\label{b2_set}
\end{align}
respectively.



From Lemma \ref{vector_set} we have the following bounds on the probability that the minimum distance condition is violated for the set \eqref{g_set}:
\begin{align}
    \nonumber \bbP\biggl(\exists (f,f') \in [F] \times [F], f\neq f': \norm{\sff - \sfprime}_0 < \frac{r-1}{20} \biggr) \\  \leq \exp\left[\log(\frac{{F}^2}{2})  - \frac{(r-1)}{2}\left(1 - \frac{2}{20}\right)^2\right] \label{probability_vector}.
\end{align}
Likewise, from Corollary \ref{matrix_set}, we have the following bounds on the probability that the minimum distance conditions are violated for the sets \eqref{b1_set} and \eqref{b2_set}, respectively: 
\begin{align}
    \nonumber &\bbP\biggl(\exists (p_1,p_1') \in [P_1] \times [P_1], p_1\neq p_1' :\norm{\Spbone - \Spboneprime}_0 \\ 
    &\nonumber \qquad <\frac{(m_1-1)r}{20} \biggr)\\ 
    & \qquad \leq \exp\left[\log\bigl(\frac{{P_1}^2}{2}\bigr) - \frac{(m_1-1)r}{2}\left(1 - \frac{2}{20}\right)^2\right],\label{probability_matrixb1}
\end{align}
and
\begin{align}
    \nonumber &\bbP\biggl(\exists (p_2,p_2') \in [P_2] \times [P_2], p_2\neq p_2' :\norm{\Spbtwo - \Spbtwoprime}_0 \\ 
    &\nonumber \qquad <\frac{(m_2-1)r}{20} \biggr) \\ 
    &\qquad \leq \exp\left[\log\bigl(\frac{{P_2}^2}{2}\bigr) - \frac{(m_2-1)r}{2}\left(1 - \frac{2}{20}\right)^2\right].\label{probability_matrixb2}
\end{align}

We require the set of coefficient matrices $\cB_L$ from the sets in \eqref{g_set}, \eqref{b1_set} and \eqref{b2_set} to exist simultaneously. Hence, using a union bound on 
\eqref{probability_vector}, \eqref{probability_matrixb1} and \eqref{probability_matrixb1}, we can choose parameters to guarantee the existence of a construction.
This is satisfied if the following conditions on the cardinalities $F$, $P_1$ and $P_2$ hold:
\begin{align}
    0< F < {\frac{\log_2(e)}{4}\left(1 - \frac{1}{10}\right)^2(r-1) - \frac{1}{2} \log_2(\frac{3}{2})} \label{F},
\end{align}
\begin{align}
    0<P_1<2^{\frac{\log_2(e)}{4}\left(1 - \frac{1}{10}\right)^2(m_1-1)r - \frac{1}{2} \log_2(\frac{3}{2})} \label{P_1},
\end{align}
and 
\begin{align}
    0<P_2<2^{\frac{\log_2(e)}{4}\left(1 - \frac{1}{10}\right)^2(m_2-1)r - \frac{1}{2} \log_2(\frac{3}{2})} \label{P_2}.
\end{align}
Note that \eqref{F}, \eqref{P_1} and \eqref{P_2} are sufficient conditions for the simultaneous existence of sets in \eqref{g_set}, \eqref{b1_set} and \eqref{b2_set}, such that the minimum distance condition between any two elements in each set is satisfied.

We proceed with the following steps in order to construct the final set $\cB_L$ of coefficient matrices. Without loss of generality, we assume that the energy of any $\Bvl$ is upper bounded by $d^2$. We will construct diagonal matrices $\bG_f$, and matrices with orthonormal columns, namely $\Bpone$ and $\Bptwo$, all of which will be used to construct each $\Bvl \in \cB_L$. In other words, due to our LR model, any matrix $\Bvl$ will have a rank-$r$ singular value decomposition. Thus, a bound on the matrix norm of $\Bvl$ gives a bound on the norm of the matrix of singular values. 

Firstly, we construct vectors $\gfone \in \R^r$ for $f \in [F]$, using $\bQ$ and $\sff$, as follows:
\begin{align}
    \gfone = \bQ \begin{bmatrix} \sqrt{\frac{1}{r-1}} \\ \sff \end{bmatrix}, \forall f \in [F].\label{gfone}
\end{align}
From \eqref{gfone}, since $\norm{\sff}^2 = 1$ we have:
\begin{align}
    \nonumber \norm{\gfone}_2^2 &= \norm{\bQ \begin{bmatrix} \sqrt{\frac{1}{r-1}} \\ \sff \end{bmatrix}}_2^2 = 
     \frac{r}{r-1}.
\end{align}
Similarly, we construct matrices $\Boneone \in \R^{m_1 \times r}$, for $p_1 \in [P_1]$ and $\Btwoone \in \R^{m_2 \times r}$, for $p_2 \in [P_2]$, respectively. Define $\boneone$ as the $j^{th}$ column of $\Boneone$, and $\bonetwo$ as the $j^{th}$ column of $\Btwoone$. Let the columns be constructed as follows:
\begin{align}
    \boneone = \bU_{1,j} \begin{bmatrix} 1 \\ \spbone \end{bmatrix}, \forall p_1 \in [P_1],\label{boneone}
\end{align}
and 
\begin{align}
    \btwoone = \bU_{2,j} \begin{bmatrix} 1 \\ \spbtwo \end{bmatrix}, \forall p_2 \in [P_2].\label{btwoone}
\end{align}
From \eqref{boneone} and \eqref{btwoone}, we have:
\begin{align}
    \nonumber \norm{\boneone}_2^2 =  \norm{\btwoone}_2^2 &= \norm{\begin{bmatrix} 1 \\ \spbone \end{bmatrix}}_2^2 = 
    \frac{r+1}{r}.
\end{align}
Secondly, we construct an $r$-sparse vector $\gftwo \in \R^{r^2}$, element-wise, from $\gfone$. Define $\gftwoi$ as the $i$th element of $\gftwo$, where $i \in [r^2]$ and use the following construction:
\begin{align}
    \gftwoi = \begin{cases} 
                |\gfonei| & i = i' +r(i'-1), i' = \{1,
                \dots,r\} \\
                0 & \textrm{otherwise}
                \end{cases},
\end{align}
and we note that 
\begin{align}
    \norm{\gftwo}_2^2 = \norm{\gfone}_2^2 = \frac{r}{r-1}. 
\end{align}
We also construct matrices $\Bonetwo \in \R^{m_1 \times r}$, for $p_1 \in [P_1]$ and $\Btwotwo \in \R^{m_2 \times r}$, for $p_2 \in [P_2]$. We show the construction of $\Bonetwo$ only: the construction of $\Btwotwo$ follows the same procedure. Define $\bonetwo \in \R^{m_1}$ as the $j^{th}$ column of $\Bonetwo$, for $j \in [r]$.  We set
\begin{align}
    \mathbf{b}_{1,p_1,1}^{(2)} = \frac{\mathbf{b}_{1,p_1,1}^{(1)}}{\norm{\mathbf{b}_{1,p_1,1}^{(1)}}_2},\label{gs_step1}
\end{align}
and define
\begin{align}
    \mathbf{a}_{j+1} \triangleq &\mathbf{b}_{1,p_1,j+1}^{(1)}  - \sum\limits_{j' = 1}^{j}  \langle\mathbf{b}_{1,p_1,j+1}^{(1)}, \mathbf{b}_{1,p_1,j'}^{(2)}\rangle \mathbf{b}_{1,p_1,j'}^{(2)}\label{gs_step2},
\end{align}
and
\begin{align}
    \mathbf{b}_{1,p_1,j+1}^{(2)} \triangleq \frac{\mathbf{a}_{j+1}}{\norm{\mathbf{a}_{j+1}}_2}\label{gs_step3},
\end{align}
for $j\in[r-1]$. 

The steps in \eqref{gs_step1}, \eqref{gs_step2} and \eqref{gs_step3} constitute the well-known Gram-Schmidt Process. Thus, set of vectors $\bonetwo$, for $j \in [r-1]$, $p_1 \in [P_1]$ are orthonormal, i.e, $\norm{\bonetwo}_2^2 = 1$ and $\bonetwo \perp \mathbf{b}_{1,p_1,j'}^{(2)}$, for any two distinct $j,j' \in [r]$. Consequently, $\left(\Btwotwo\right)^T\left(\Btwotwo\right) = \mathbf{I}_r$.

Finally, we define the vector $\gf$ and matrices $\Bpone$ and $\Bptwo$ as:
\begin{align}
    \gf = \frac{\varepsilon}{r} \gftwo, \forall f \in [F],
\end{align}
and
\begin{align}
    \Bpone = \Bonetwo, ~ \Bptwo = \Btwotwo,
\end{align}
respectively, for some positive number $\varepsilon$.

We also define the diagonal matrix $\bG_{f} \in \R^{r \times r}$, where $\tvec(\bG_{f} ) = \gf$. 

Now, by designating 
\begin{align}
    \cL \triangleq \left\{ (f, p_1, p_2): f \in [F], p_1 \in [P_1], p_2 \in [P_2] \right\},
\end{align}    
as the set of possible tuples $(f, p_1, p_2)$, we have
\begin{align}
    L = |\cL| &
    \overset{(a)}{<} 2^{\left[\frac{\log_2(e)}{4}\left(c_1 \left(r(m_1 +m_2 -2) +  (r-1)\right)\right) - (\frac{3}{2})\log_2\left(\frac{3}{2}\right)\right]},
\end{align}
where $(a)$ follows from \eqref{F}, \eqref{P_1} and \eqref{P_2}. We define the set of coefficient matrices, $\cB_L$ as, 
\begin{align}
    \cB_L \triangleq \biggl\{ \Bvl = \Bpone^T \bG_{f}  \Bptwo, l \in [L], f \in [F], \\ p_1 \in [P_1], p_2 \in [P_2]\biggr\},
\end{align}
and we restrict $\varepsilon$ such that 
\begin{align}
    \sqrt{\frac{8(r-1)}{r}} < \varepsilon < d\sqrt{\frac{r-1}{r}}, \label{eps_cond}
\end{align}
in order to guarantee $\norm{\bG}_2^2 < \frac{d^2}{r^2}$. We make the final note that, due to the Kronecker product, we can express $\tvec(\Bvl)$ as:
\begin{align}
    \tvec(\Bvl) = (\Bpone \otimes \Bptwo) \gf.
\end{align}

We have the following remaining tasks at hand: 1) We must show that they energy of any $\Bvl$ is less than $d^2$. 2) We must derive upper and lower bounds on the distance between any two distinct $\Bvl, \Bvlprime \in B_L$ $\left(\norm{\Bvl -\Bvlprime}_F^2\right)$. 

We begin by showing $\norm{\Bvl}_F^2 < d^2$:
\begin{align}
    \nonumber \norm{\Bvl}_F^2
        & = \norm{\left(\Bptwo \otimes \Bpone\right)\gf}_2^2\\
        &\nonumber \leq \norm{\Bptwo\otimes \Bpone}_F^2 \norm{\gf}_2^2\\
        &\nonumber \overset{(b)}{=} \norm{\Bpone}_F^2 \norm{\Bptwo}_F^2 \norm{\gf}_2^2 = \frac{r\varepsilon^2}{r-1} \overset{(c)}{<} d^2,
\end{align}
where $(b)$ follows from the fact that the matrix norm of the Kronecker product is the product of the matrix norms, and $(c)$ holds due to \eqref{eps_cond}. 

We proceed with deriving lower and upper bounds on $\norm{\Bvl -\Bvlprime}_F^2$, for any two distinct $\Bvl, \Bvlprime \in B_L$. For finding lower bounds on $\norm{\Bvl -\Bvlprime}_F^2$, it can be shown that the closest pair $\Bvl, \Bvlprime$ occurs for $p_1 = p_1'$,  $p_2 = p_2'$ and $f \neq f'$. Thus we have the following:
\begin{align}
        \nonumber \norm{\Bvl - \Bvlprime}_F^2 &\nonumber= \norm{\left(\Bptwo \otimes \Bpone \right)\gf -\left(\Bptwoprime \otimes \Bponeprime \right)\gfprime}_2^2\\
        & \nonumber \geq \norm{\left(\Bptwo \otimes \Bpone\right) \left(\gf - \gfprime\right)}_2^2\\
        &\nonumber \overset{(d)}{=} \frac{\varepsilon^2}{r^2}\norm{\gftwo - \gftwoprime}_2^2\\
        & \nonumber
         >\frac{r\varepsilon^2}{r-1}\frac{4}{20} \label{dist_LB}
\end{align} 
where $(d)$ follows from the fact that the Kronecker product of orthogonal bases is an orthogonal basis.   

Finally, for finding upper bounds on $\norm{\Bvl - \Bvlprime}_F^2$, we have:
\begin{align}
    \nonumber &\norm{\Bvl - \Bvlprime}_F^2 \\ &\nonumber \overset{(e)}{\leq} \left[\norm{\left(\Bptwo \otimes \Bpone\right) \gf}_F +\norm{\left(\Bptwoprime \otimes \Bponeprime \right) \gfprime}_F\right]^2\\
    &\nonumber \leq \left[\norm{\Bptwo \otimes \Bpone}_F\norm{\gf}_2 +\norm{\Bptwoprime \otimes \Bponeprime}_F\norm{ \gfprime}_2\right]^2\\
    & = \left[ 2\norm{\Bpone}_F \norm{\Bptwo}_F \norm{\gf}_2 \right]^2 \\
    & \leq 4\frac{r\varepsilon^2}{r-1} \label{dist_UB}
\end{align}
where $(e)$ follows from the triangle inequality.  
\end{IEEEproof}

\begin{IEEEproof}[Proof of Lemma 3]
Consider the set $\cB_L$ defined in Lemma 2, where the bounds in \eqref{dist_LB} and \eqref{dist_UB} hold. Consider the matrix LR model in \eqref{gen_matrix_model}. For $n$ i.i.d samples, consider covariate matrices $\bX_i \in \R^{m_1 \times m_2}, \forall i \in [n]$, where $\tvec(\bX_i) \sim \cN(\mathbf{0}, \sigma^2\mathbf{I}_{m_1m_2})$. According to \eqref{gen_matrix_model}, observations $y_i$ follow a Bernoulli distribution when conditioned on $\bX_i$, $\forall i \in [n]$. Consider $\by$ and $\tnsrX^c$ defined in Section \ref{model_prob_form}, and define $\cmi{\by}{l}{\tnsrX^c}$ as the mutual information between observations $\by$ and index $l$ conditioned on side-information $\tnsrX^c$. From \cite{cover2012elements, wainwright2009information}, we have
\begin{align}
    \cmi{\by}{l}{\tnsrX^c} \leq \frac{1}{L^2} \sum_{l,l'}\E_{\tnsrX^c} D_{KL}(f_l(\by|\tnsrX^c)||f_{l'}(\by|\tnsrX^c) \label{cmi},
\end{align}
where $D_{KL}(f_l(\by|\tnsrX^c)||f_{l'}(\by|\tnsrX^c)$ is the Kullback-Leibler (KL) divergence of  probability distribution $f_l(\by|\tnsrX^c)$ of $\by$ given $\tnsrX^c$ for some $\Bvl \in \cB_L$. Denote $\sigma_l \triangleq \frac{1}{1+ \exp{-\langle\Bvl, \bX_i\rangle}}$, and  $\sigma_{l'} \triangleq \frac{1}{1+ \exp^{-\langle\Bvlprime, \bX_i\rangle}}$, We evaluate the KL divergence as follows:
\begin{align}
    \nonumber D&_{KL}(f_l(\by|\tnsrX^c)||f_{l'}(\by|\tnsrX^c)) \\
    \nonumber &=\sum\limits_{i \in [n]} \sigma_l \log\left(\frac{\sigma_l}{\sigma_{l'}}\right) +(1-\sigma_l)\log\left(\frac{1- \sigma_l}{1-\sigma_{l'}}\right).\\
    \nonumber & = \sum\limits_{i \in [n]} \sigma_l\langle\Bvl, \bX_i\rangle - \sigma_{l}\langle\Bvlprime, \bX_i\rangle - \langle\Bvl, \bX_i\rangle + \langle\Bvlprime, \bX_i\rangle \\
    & + \log(1 + \exp^{-\langle\Bvl, \bX_i\rangle}) - \log(1 + \exp^{-\langle\Bvlprime, \bX_i\rangle}) \label{KL}. 
\end{align}
Now, considering the distribution on covariates $\bX_i, \forall i \in [n]$, we take the expectation of \eqref{KL} with respect to the side-information $\tnsrX^c$. We have $\E_{\tnsrX^c}(\log(1 + \exp^{-\langle\Bvl, \bX_i\rangle})) = \E_{\tnsrX^c}(\log(1 + \exp^{-\langle\Bvlprime, \bX_i\rangle}))$, and $\E_{\tnsrX^c}(\langle\Bvl, \bX_i\rangle) = \E_{\tnsrX^c}(\langle\Bvlprime, \bX_i\rangle) = 0$. We are left with:
\begin{align}
    \nonumber \E_{\tnsrX^c} D_{KL}&(f_l(\by|\tnsrX^c)||f_{l'}(\by|\tnsrX^c)\\
    \nonumber &= \sum\limits_{i \in[n]} \E_{\bX} \bigl[\sigma_l\cdot\langle \bX_i ,\Bvl\rangle - \sigma_l\langle \bX_i, \Bvlprime\rangle\bigr]\\
    \nonumber &=\sum\limits_{i \in[n]} \E_{\bX}\bigl[\sigma_l\cdot\langle\bX_i,\Bvl-\Bvlprime\rangle \bigr]\\
    \nonumber &=\sum\limits_{i \in[n]} \E_{\bX}\biggl[\frac{ \langle \bX_i, \Bvl - \Bvlprime\rangle}{1+\exp^{-\langle\bX_i,\Bvl\rangle}}\biggr]\\
    &\leq n \E_{\bX} \biggl[ \left| \langle \tvec(\bX_i), \tvec(\Bvl - \Bvlprime)\rangle\right| \biggr] \label{exp_KL}.
\end{align}
Define the random variable $\widetilde{X} \triangleq \langle \bX_i, \Bvl-\Bvlprime \rangle$. $\widetilde{X}$ is a normally distributed random variable with mean  $\mu_{\{\widetilde{X}\}} = 0$ and variance $\sigma_{\widetilde{X}}^2 =\sigma^2 \tvec(\Bvl - \Bvlprime)^T \mathbf{I}_{m_1m_2}\tvec(\Bvl - \Bvlprime)$. Now, define the half-Normal random variable $\bar{X} \triangleq |\widetilde{X}|$. $\bar{X}$ is a half-Normal distribution with mean:
\begin{align}
     \E_{\bar{X}}[\bar{X}] &= \sigma \sqrt{\tvec(\Bvl -\Bvlprime)^T\tvec(\Bvl -\Bvlprime)}\sqrt{\frac{2}{\pi}} \notag \\
     &= \sigma \norm{\Bvl -\Bvlprime}_F\sqrt{\frac{2}{\pi}}.\label{expectation}
\end{align}
Plugging in \eqref{expectation} and \eqref{exp_KL} into \eqref{cmi} gives us,
\begin{align}
    \nonumber \cmi{\by}{l}{\tnsrX^c} &\leq n \sigma \norm{\Bvl -\Bvlprime}_F\sqrt{\frac{2}{\pi}}
    \overset{(f)}{\leq} n \sigma \sqrt{\frac{4r}{r-1}}\sqrt{\frac{2}{\pi}} \varepsilon,
\end{align}
where $(f)$ follows from \eqref{dist_UB}. 
\end{IEEEproof}

\balance


\begin{thebibliography}{10}
\providecommand{\url}[1]{#1}
\csname url@samestyle\endcsname
\providecommand{\newblock}{\relax}
\providecommand{\bibinfo}[2]{#2}
\providecommand{\BIBentrySTDinterwordspacing}{\spaceskip=0pt\relax}
\providecommand{\BIBentryALTinterwordstretchfactor}{4}
\providecommand{\BIBentryALTinterwordspacing}{\spaceskip=\fontdimen2\font plus
\BIBentryALTinterwordstretchfactor\fontdimen3\font minus
  \fontdimen4\font\relax}
\providecommand{\BIBforeignlanguage}[2]{{%
\expandafter\ifx\csname l@#1\endcsname\relax
\typeout{** WARNING: IEEEtran.bst: No hyphenation pattern has been}%
\typeout{** loaded for the language `#1'. Using the pattern for}%
\typeout{** the default language instead.}%
\else
\language=\csname l@#1\endcsname
\fi
#2}}
\providecommand{\BIBdecl}{\relax}
\BIBdecl

\bibitem{wright1995logistic}
R.~E. Wright, ``Logistic regression.'' in \emph{Reading and Understanding
  Multivariate Statistics}.\hskip 1em plus 0.5em minus 0.4em\relax American
  Psychological Association, 1995, pp. 217--244.

\bibitem{dumas2019computational}
J.~P. Dumas, M.~A. Lodhi, B.~A. Taki, W.~U. Bajwa, and M.~C. Pierce,
  ``Computational endoscopy—a framework for improving spatial resolution in
  fiber bundle imaging,'' \emph{Optics Letters}, vol.~44, no.~16, pp.
  3968--3971, 2019.

\bibitem{lodhi2020learning}
M.~A. Lodhi and W.~U. Bajwa, ``Learning product graphs underlying smooth graph
  signals,'' \emph{arXiv preprint arXiv:2002.11277}, 2020.

\bibitem{hung2013matrix}
H.~Hung and C.-C. Wang, ``Matrix variate logistic regression model with
  application to {EEG} data,'' \emph{Biostatistics}, vol.~14, no.~1, pp.
  189--202, 2013.

\bibitem{zhang2018rank}
J.~Zhang and J.~Jiang, ``Rank-optimized logistic matrix regression toward
  improved matrix data classification,'' \emph{Neural Computation}, vol.~30,
  no.~2, pp. 505--525, 2018.

\bibitem{barnes2019minimax}
L.~P. Barnes and A.~Ozgur, ``Minimax bounds for distributed logistic
  regression,'' \emph{arXiv preprint arXiv:1910.01625}, 2019.

\bibitem{zhang2020islet}
A.~R. Zhang, Y.~Luo, G.~Raskutti, and M.~Yuan, ``{ISLET}: Fast and optimal
  low-rank tensor regression via importance sketching,'' \emph{SIAM J. on
  Mathematics of Data Science}, vol.~2, no.~2, pp. 444--479, 2020.

\bibitem{raskutti2019convex}
G.~Raskutti, M.~Yuan, and H.~Chen, ``Convex regularization for high-dimensional
  multiresponse tensor regression,'' \emph{The Annals of Statstics}, vol.~47,
  no.~3, pp. 1554--1584, 2019.

\bibitem{foster2018logistic}
D.~J. Foster, S.~Kale, H.~Luo, M.~Mohri, and K.~Sridharan, ``Logistic
  regression: The importance of being improper,'' in \emph{Proc. Conf. On
  Learning Theory (PMLR)}, 2018, pp. 167--208.

\bibitem{abramovich2018high}
F.~Abramovich and V.~Grinshtein, ``High-dimensional classification by sparse
  logistic regression,'' \emph{IEEE Trans. on Information Theory}, vol.~65,
  no.~5, pp. 3068--3079, 2018.

\bibitem{shi2014sparse}
J.~V. Shi, Y.~Xu, and R.~G. Baraniuk, ``Sparse bilinear logistic regression,''
  \emph{arXiv preprint arXiv:1404.4104}, 2014.

\bibitem{an2020logistic}
B.~An and B.~Zhang, ``Logistic regression with image covariates via the
  combination of $\ell_1$ and {S}obolev regularizations,'' \emph{PLOS One},
  vol.~15, no.~6, p. e0234975, 2020.

\bibitem{berthet2020statistical}
Q.~Berthet and N.~Baldin, ``Statistical and computational rates in graph
  logistic regression,'' in \emph{Proc. Int. Conf. on Artificial Intelligence
  and Statistics (PMLR)}, 2020, pp. 2719--2730.

\bibitem{jung2016minimax}
A.~Jung, Y.~C. Eldar, and N.~G{\"o}rtz, ``On the minimax risk of dictionary
  learning,'' \emph{IEEE Trans. on Inf. Theory}, vol.~62, no.~3, pp.
  1501--1515, 2016.

\bibitem{shakeri2016minimax}
Z.~Shakeri, W.~U. Bajwa, and A.~D. Sarwate, ``Minimax lower bounds for
  kronecker-structured dictionary learning,'' in \emph{Proc. 2016 IEEE Int.
  Symp. on Inf. Theory (ISIT)}.\hskip 1em plus 0.5em minus 0.4em\relax IEEE,
  2016, pp. 1148--1152.

\bibitem{yu1997assouad}
B.~Yu, ``Assouad, {Fano}, and {Le Cam},'' in \emph{Festschrift for Lucien Le
  Cam}.\hskip 1em plus 0.5em minus 0.4em\relax Springer, 1997, pp. 423--435.

\bibitem{kolda2009tensor}
T.~G. Kolda and B.~W. Bader, ``Tensor decompositions and applications,''
  \emph{{SIAM} Review}, vol.~51, no.~3, pp. 455--500, 2009.

\bibitem{negahban2012restricted}
S.~Negahban and M.~J. Wainwright, ``Restricted strong convexity and weighted
  matrix completion: Optimal bounds with noise,'' \emph{The J. of Machine
  Learning Res. (JMLR)}, vol.~13, pp. 1665--1697, 2012.

\bibitem{khas1979lower}
R.~Z. Khas'minskii, ``A lower bound on the risks of non-parametric estimates of
  densities in the uniform metric,'' \emph{Theory of Probability \& Its
  Applications}, vol.~23, no.~4, pp. 794--798, 1979.

\bibitem{cover2012elements}
T.~M. Cover and J.~A. Thomas, \emph{Elements of Information Theory}.\hskip 1em
  plus 0.5em minus 0.4em\relax John Wiley \& Sons, 2012.

\bibitem{dubhashi2009concentration}
D.~P. Dubhashi and A.~Panconesi, \emph{Concentration of Measure for the
  Analysis of Randomized Algorithms}.\hskip 1em plus 0.5em minus 0.4em\relax
  Cambridge University Press, 2009.

\bibitem{shakeri2018minimax}
Z.~Shakeri, W.~U. Bajwa, and A.~D. Sarwate, ``Minimax lower bounds on
  dictionary learning for tensor data,'' \emph{IEEE Trans. on Inf. Theory},
  vol.~64, no.~4, pp. 2706--2726, 2018.

\bibitem{wainwright2009information}
M.~J. Wainwright, ``Information-theoretic limits on sparsity recovery in the
  high-dimensional and noisy setting,'' \emph{IEEE Trans. on Inf. Theory},
  vol.~55, no.~12, pp. 5728--5741, 2009.

\end{thebibliography}
\end{document}